%% file: main.tex
\PassOptionsToPackage{table}{xcolor}
\documentclass[10pt,twocolumn,letterpaper]{article}

\usepackage{cvpr}              % To produce the CAMERA-READY version
\input{include/packages}
\input{include/cmds}

\input{include/shorthands}

\definecolor{cvprblue}{rgb}{0.21,0.49,0.74}
\usepackage[pagebackref,breaklinks,colorlinks,allcolors=cvprblue]{hyperref}

\def\paperID{10613} % *** Enter the Paper ID here
\def\confName{CVPR}
\def\confYear{2025}

\title{Efficient Depth Estimation for Unstable Stereo Camera Systems on AR Glasses}
\author{Yongfan Liu ~~~~~ Hyoukjun Kwon\\
University of California, Irvine \\
Irvine, CA, USA \\
{\tt\small \{yongfal, hyoukjun.kwon\}@uci.edu}
}

\begin{document}
\maketitle
\input{sections/00_Abstract}
\input{sections/01_Introduction}

\input{sections/02_Rel_work}

\input{sections/03_MultiHeadDepth}
\input{sections/04_HomoDepth}

\input{sections/05_Evaluation}

\input{sections/06_Conclusion}

{
    \small
    \bibliographystyle{ieeenat_fullname}
    \bibliography{main}
}


\input{sections/10_Appendix}

\end{document}

%% file: include/packages.tex
\usepackage{amsmath,amssymb,amsfonts,amsthm}
\usepackage[italic]{mathastext}
\usepackage{algorithm}
\usepackage{algpseudocode}
\usepackage[toc,page]{appendix}

\usepackage{caption}
\usepackage{tabularx}
\usepackage{tabularray}
\usepackage{multirow}

\usepackage{booktabs}
\usepackage[table]{xcolor}
\usepackage{comment}

%% file: include/cmds.tex
\usepackage{subcaption}

%% file: include/shorthands.tex
\newcommand{\argos}{\textsc{Argos}\xspace}
\newcommand{\mulh}{\textsc{MultiHeadDepth}\xspace}
\newcommand{\homo}{\textsc{HomoDepth}\xspace}
\newcommand{\mulhc}{\textsc{MultiheadCostVolume}\xspace}

%% file: sections/00_Abstract.tex
\begin{abstract}

\vspace{-6mm}
Stereo depth estimation is a fundamental component in augmented reality (AR), which requires low latency for real-time processing. However, preprocessing such as rectification and non-ML computations such as cost volume require significant amount of latency exceeding that of an ML model itself, which hinders the real-time processing required by AR. 
Therefore, we develop alternative approaches to the rectification and cost volume that consider ML acceleration (GPU and NPUs) in recent hardware. For pre-processing, we eliminate it by introducing homography matrix prediction network with a rectification positional encoding (RPE), which delivers both low latency and robustness to unrectified images. For cost volume, we replace it with a group-pointwise convolution-based operator and approximation of cosine similarity based on layernorm and dot product.

\begin{comment}   
However, traditional depth estimation models often rely on time-consuming preprocessing steps such as rectification to achieve high accuracy. Also, non-standard ML operator-based algorithms such as cost volume also require significant latency, which is aggravated on compute resource-constrained mobile platforms. Therefore, we develop hardware-friendly alternatives to the costly cost volume and preprocessing and design two new models based on them, MultiHeadDepth and HomoDepth.
Our approach for cost volume is replacing it with a new group-pointwise convolution-based operator and approximation of cosine similarity based on layernorm and dot product. For online stereo rectification (preprocessing), we introduce homography matrix prediction network with a rectification positional encoding (RPE), which delivers both low latency and robustness to unrectified images, which eliminates the needs for preprocessing.
\end{comment}

Based on our approaches, we develop MultiHeadDepth (replacing cost volume) and HomoDepth (MultiHeadDepth + removing pre-processing) models. MultiHeadDepth provides 11.8-30.3\% improvements in accuracy and 22.9-25.2\% reduction in latency compared to a state-of-the-art depth estimation model for AR glasses from industry. HomoDepth, which can directly process unrectified images, reduces the end-to-end latency by 44.5\%. We also introduce a multi-task learning method to handle misaligned stereo inputs on HomoDepth, which reduces the AbsRel error by 10.0-24.3\%. The overall results demonstrate the efficacy of our approaches, which not only reduce the inference latency but also improve the model performance. Our code is available at \url{https://github.com/UCI-ISA-Lab/MultiHeadDepth-HomoDepth}

\begin{comment}
    Our MultiHeadDepth, which includes optimized cost volume, provides 11.8-30.3\% improvements in accuracy and 22.9-25.2\% reduction in latency compared to a state-of-the-art depth estimation model for AR glasses from industry. Our HomoDepth, which includes optimized preprocessing (homography + RPE) upon MultiHeadDepth, can process unrectified images and reduce the end-to-end latency by 44.5\%. We adopt a multi-task learning framework to handle misaligned stereo inputs on HomoDepth, which reduces the AbsRel error by 10.0-24.3\%. The results demonstrate the efficacy of our approaches in achieving both high model performance with low latency, which makes a step forward toward practical depth estimation on future AR devices. Codes are available at \url{https://github.com/UCI-ISA-Lab/MultiHeadDepth-HomoDepth}
\end{comment}

\end{abstract}

%% file: sections/01_Introduction.tex
\section{Introduction}
\label{sec:intro}

%\begin{figure*}[t]
%        \centering
        %\includegraphics[height=0.18\textheight]{figures/head.png}
        %\vspace{-3mm}
        %\caption{\small \textbf{Overview} A detailed latency breakdown information is shown as \cref{fig:late_breakdown}}
        %\label{fig:head}
%\end{figure*}

Depth estimation serves as a foundational component in augmented and virtual reality (AR/VR)~\cite{kwon2023xrbench} with many downstream algorithms, which include novel-view rendering~\cite{Tiefenrausch, wang2023practical}, occlusion reasoning \cite{li2021roboticocclusionreasoningefficient}, world locking for AR object placement \cite{10316484}, and determining the scale of AR objects \cite{10.1007/978-3-030-23528-4_46}.
In the AR domain, stereo depth estimation is often deployed~\cite{wang2023practical} rather than mono depth estimation due to its superior accuracy and the ease of deploying stereo cameras on the each side of AR glasses frames.
Due to the realtime nature of AR applications~\cite{kwon2023xrbench}, one key objective for depth estimation models for AR is low latency, in addition to good performance. However, since AR glasses are in a compute resource-constrained wearable form factor, enabling desired latency (less than 100 ms on device) is not trivial.

One challenge towards the latency optimization originates from the preprocessing, which is a indispensable step for achieving high model performance in practical applications.
Examples include camera calibration and rectification using camera intrinsic and extrinsic. In addition to preprocessing, traditional algorithms embedded into depth estimation models (e.g, cost volume~\cite{GAN2021118, wang2023practical}) also imposes another challenge to the latency optimization. We present their significance in latency in \cref{fig:late_breakdown}, which shows that preprocessing accounts for 30.2\% and cost volume accounts for 29.3\%  of the total latency of a state-of-the-art (SOTA) model, \argos~\cite{wang2023practical}.
Note that preprocessing latency can be more dominant when camera intrinsic and extrinsic parameters are unknown, which results in 200 to 2000 ms latency to solve for the extrinsic parameters and subsequently process the images \cite{Kumar2010StereoRO, MobiDepth}. This aggravates the long preprocessing latency challenge further.

Therefore, to reduce depth estimation model latency for AR glasses, we propose new methodologies that significantly reduce the preprocessing (online stereo rectification not required) and cost volume latency, as shown in~\cref{fig:late_breakdown}. Our approach for the cost volume is (1) to replace traditional algorithm with group-pointwise convolutions, which are highly optimized in hardware and compilers and (2) adopt an efficient approximation of cosine similarity using layernorm and dot product. For preprocessing (stereo matching), we adopt homography matrix-based approximation and estimate homography using a head attached to the depth estimation model, which allows to utilize homography matrix-based approximation with dynamically varying extrinsic parameters in unstable AR glasses platform or no access to camera extrinsic parameters.

Since our methodologies are complementary to existing depth estimation models, we augment our methodologies on the \argos ~\cite{wang2023practical} and develop two new models, \mulh and \homo. As presented in~\cref{fig:late_breakdown}, \mulh focuses on cost volume optimization, which reduces inference latency by 25\% compared to the original cost volume. \homo targets scenarios that require stereo matching preprocessing, which can directly accept unrectified images as input, eliminating the needs for preprocessing. \homo not only provides high-quality outputs on unrectified images as presented in~\cref{tab:ablation}, but also significantly reduces the end-to-end latency by 43\%. Our evaluation includes ADT~\cite{ADT} dataset collected by real research AR glasses by Meta, Aria~\cite{aria}, which demonstrates the effectiveness of our approach on realistic AR glasses platform.

We summarize our contributions as follows:

\begin{itemize}
  {\item We develop \mulhc block that replaces cost volume in a previous depth estimation model. Our new block provides significantly lower latency as well as higher accuracy.}

  {\item We introduce the homography of stereo images to reveal their position relationship to enable to accept unrectified images without preprocessing. We merge a small homography estimation head within the depth estimation network, which significantly reduces the latency compared to the preprocessing-based approach.}
 
  {\item We augment our homography estimation head with 2D rectification position encoding, which helps translate relative positional information from homograpghy matrix to the 2D rectification position encoding format. It enables the neural network to effectively understand relative position information, which plays a key role in eliminating the need for rectification preprocessing.}
  
\end{itemize}

\input{figures/late_breakdown}

%% file: figures/late_breakdown.tex
\begin{figure}[t]
        \centering
        \includegraphics[width=0.9\linewidth]{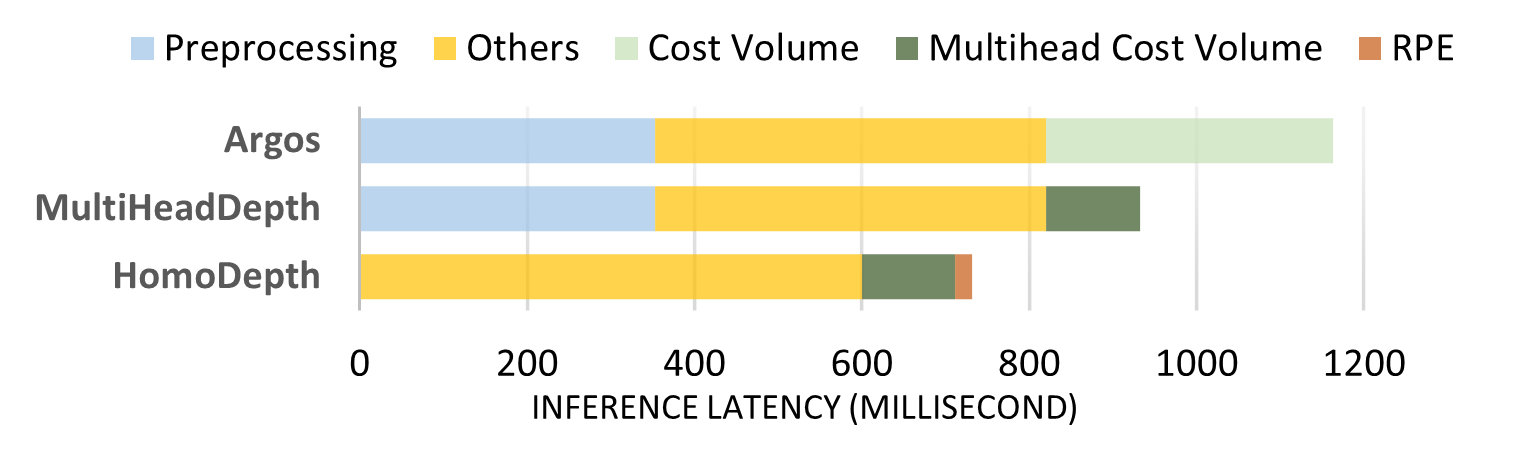}
        \vspace{-4mm}
        \caption{\small \textbf{The latency breakdown analysis of a SOTA model \argos~\cite{wang2023practical} and ours, \mulh \& \homo} on Intel i7-12700H laptop CPU. The "RPE" refers to the 2D rectification position encoding process. "Others" refers to all the other parts of the neural network excluding cost volume blocks, such as Conv, Norm, FC, and ReLU6.}
        \vspace{-6mm}
        \label{fig:late_breakdown}
    \end{figure}

%% file: sections/02_Rel_work.tex
\section{Related Work}
\label{sec:background}

\paragraph{Stereo Depth Estimation.} Fast depth estimation on mobile devices is a challenging task. Multiple previous works focused on delivering both model performance and efficiency targeting mobile platforms.

%If we set a more lenient benchmark, taking the latest Aria glasses \cite{aria} as an example, the frame rate of their cameras is 10 fps. This means the inference latency for the entire depth estimation pipeline should be under 100 ms.
%Additionally, AR scenarios require dense depth estimation rather than sparse depth estimation of select points. The output demands a dense depth map, significantly increasing computational resource requirements.

\textsc{MobileStereoNet}\xspace \cite{MobileStereoNet} is a stereo depth estimation model optimized for mobile devices, featuring a lightweight network design and algorithms specially adapted for stereo matching. However, despite its minimal parameter size, the computational complexity is not small with high FLOPS (e.g. 190G FLOPS), which makes it challenging to achieve efficient performance on mobile platforms.

\argos~\cite{wang2023practical} extended \textsc{Tiefenrausch}\xspace  monocular network \cite{Tiefenrausch} to implement stereo depth estimation. \argos is optimized to minimize the model size and computational complexity. \argos consists of an encoder, cost volume, and a decoder, which requires preprocesing if the input images are distorted or not rectified. \argos was designed for practical AR applications by Meta, and it delivers SOTA performance in the stereo depth estimation on AR glasses domain.

\textsc{DynamicStereo}\xspace \cite{DynamicStereo} is a stereo depth estimation model designed for AR and mobile devices with a focus on efficiency, but \argos delivers superior accuracy and latency for just one-shot estimation.

%-------------------------------------------------------------------------
\noindent \textbf{Stereo Image Preprocessing.} The stereo image preprocessing consists of calibration and rectification. Calibration is required for a raw image captured by the camera, which removes distortion and artifacts in the raw image. %Sometimes it crops images to remove the black margins around the images.
Calibration depends on the intrinsic parameters of camera systems. Generally, the intrinsic of a camera is stable and provided by manufacturers. Calibration can be completed with just a few matrix computations, within a few dozen milliseconds.

Rectification is a major challenge in stereo vision systems. For an object in the real world, its projection onto two image planes should lie on the epipolar lines, satisfying the epipolar constraint \cite{faugeras1993three}. If the two cameras are misaligned, it is necessary to rectify the pair of images so that corresponding points lie on the same horizontal lines. 
The stereo cameras are mounted on both sides of AR glasses. The glasses may undergo significant bending ($>10^{\circ}$) because of the soft material \cite{wang2023practical}. This issue can be exacerbated by variations in users' head sizes and temperatures. 

Rectification requires the relative position information of two cameras. \textsc{MobiDepth}\xspace \cite{MobiDepth} assumed that the relative positions of the cameras had been obtained before the inference. %It can only process images based on the known position relationship. This is considered as offline rectification.
However, practically, the camera positions on AR glasses need to be determined based on a pair of arbitrarily captured images, which requires online rectification unlike \textsc{MobiDepth}\xspace's offline method. 
Karaev \etal \cite{DynamicStereo} set image saturation to a value sampled uniformly between 0 and 1.4 during the model training phase to make the model more robust for stereo images misalignment. But it lacks good interpolation for robustness. 
Wang \etal proposed a fast online rectification \cite{wang2023practical}. However, there remains a 15\% to 23\% chance of failing to rectify the stereo images.

%-------------------------------------------------------------------------

%\input{figures/homo_rect}
Every stereo-matching algorithm includes a mechanism for evaluating the similarity of image locations. Matching cost is calculated for every pixel over the range of examined disparities, and it's a commonly used measurement \cite{4270273}.
Kendall \etal \cite{Kendall_2017_ICCV} adopted deep unary features to compute the stereo matching cost by forming a cost volume. They form a cost volume of dimensionality $height \times weight \times max\_disparity \times features$ and pack it into a 4D volume. To construct the 4D volume, the algorithm iteratively move one of the images/features horizontally and compare the matching cost in each disparity, as described in \cref{fig:cv_compare_1}. The cost volume information is crucial for a neural network to understand the differences between the two feature maps.
Although proven effective, the cost volume has a disadvantage in the computational efficiency from its cosine similarity computation described below: 
\begin{equation}
    D_{cos}(\boldsymbol a, \boldsymbol b)=\boldsymbol a\cdot \boldsymbol b ~/~ |\boldsymbol a||\boldsymbol b|
    \label{eq:cos_sim}
\end{equation}
where $\boldsymbol a$ and $\boldsymbol b$ are the vector of pixels. The numerator is matrix multiplication-optimized hardware- and compiler-friendly since the dot product computation in a batch is a matrix multiplication. However, the denominator requires to compute norms pixel by pixel, which is not friendly to hardware highly optimized for matrix multiplication. Therefore, we first focus on the optimization of cost volume, which we discuss next.

%An image or feature map contains thousands to millions of pixels. For batch computation, the vector dot product in the numerator is essentially matrix multiplication, which is hardware-friendly. However, for the denominator, each element requires computing the norm before multiplication. For different $a$ and $b$, the denominator needs to be calculated separately. This means that in cosine similarity, the normalization operation from the denominator is not friendly to hardware highly optimized for GEMM operators (e.g., NPUs).

%% file: sections/03_MultiHeadDepth.tex
\section{MultiHeadDepth Model}
\label{sec:mulh}

Our approach to optimize cost volume is to replace it by a hardware-friendly approximation. We discuss our approach in detail and our solution, multi-head cost volume, that implements our approach.
%We discuss how we replace the cost volume with a hardware-friendly approximation, multi-head cost volume.
%Based our observation on the cost volume, we  optimize it by approximating cosine similarity with a set of hardware-friendly operations and introduce a new hardware-friendly block, multi-head cost volume, to replace the overall cost volume algorithm.

\begin{figure}[t]
\label{fig:cos_compare}
    \vspace{-2mm}
    \centering
    \includegraphics[width=\linewidth]{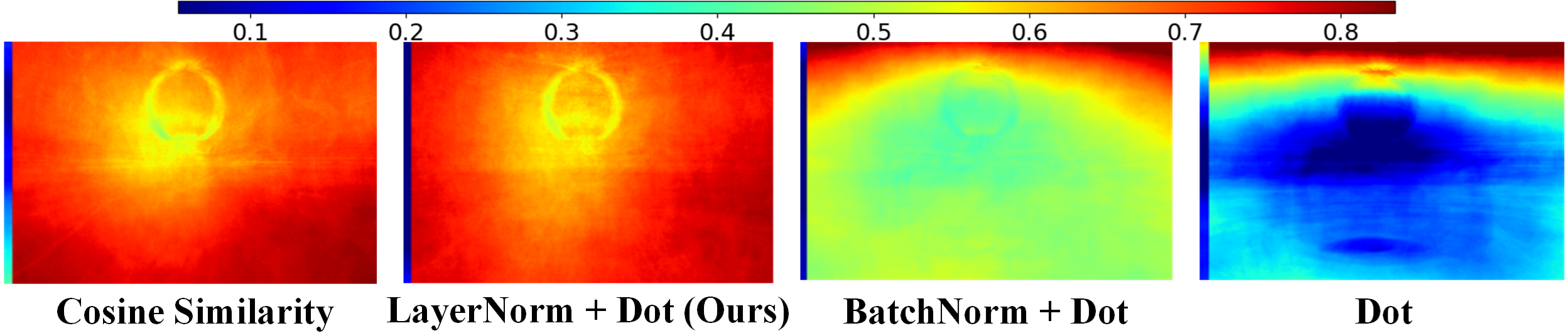}
    \vspace{-6mm}
    \caption{\small \textbf{Similarity estimation methodology Comparison.} Each represents a similarity map between left and right features after applying \textit{roll} operation (offset: 10) to the input stereo images. The maps are average results across the entire dataset of Sceneflow \cite{Sceneflow}, after rescaling to [0,1] range.
    The strips on the left side of maps are caused by \textit{roll} and they are part of the maps. }
    \vspace{-4mm}
    \label{fig:cos_compare}
\end{figure}

\subsection{Approximating Cosine Similarity}
\label{subsec:apprx_cosine}
%Continuing the discussion of cost volume block, we have shown the shortcomings of cosine similarity. Here we use 2D layer normalization and dot product to replace it.  
%The cosine similarity calculates the angle between two vectors, which indicates the similarity of the vectors. Since the denominator needs to be optimized as we discussed in~\cref{sec:background}, we focus on the denominator. 

As discussed in~\cref{sec:background}, the denominator of cosine similarity (computing two norms and multiplication of them) and the corresponding division are the prime optimization targets. The goal of those operations is to normalize individual vectors to bound the results in the [-1,1] range. Based on the observation, we approximate the normalization for individual vectors using 2D layer normalization. This approach significantly reduces the number of costly normalization computations, as discussed in \cref{subsec:compute_complexity_mhcv}. We still keep the numerator of cosine similarity as a dot product operator after the normalization. We term our approach as LND (LayerNorm combined with Dot product). LND successfully approximates the cosine similarity, as shown by the results in \cref{fig:cos_compare}.
Another benefit of the layer norm is that it provides a buffer for the cost volume, which can accept additional encoded information. That is, each normalized pixel can be merged with other encodings. For instance, it can handle positional encoding, which we discuss in \cref{sec:homo_struct}.

%Therefore, we focus on replacing the division by denominator in ~\cref{eq:cos_sim} with 2D layer normalization (layer norm) since this approach reduces the computational complexity and achieves normalization effect (i.e., the elements within a pixel vector have a mean of zero and a standard deviation of one). Maintaining the numerator (dot product) as is in the modified distance metric, the computational complexity is reduced, as discussed in~\cref{subsec:compute_complexity_mhcv}. 

% We conduct a small experiment in \cref{subsec:appendix_apprx_cosine} showing the approximation is close, also our evaluation results indicate that the metric is effective.

%we visualize the similarity values between stereo images from cosine similarity and three approximation methods including LND. 
%We input a stereo RGB image after applying the \textit{roll} with 10 offset to the right image. The 2D similarity maps, as shown in  \cref{fig:cos_compare}, are generated using Sceneflow \cite{Sceneflow}, averaged across the entire dataset, and rescaled to 0-1. The results indicate that LND approximates the cosine similarity well.

To further enhance the model performance, we also augment the LND with a weight layer after the LND, which help adjust the similarity approximation. Our evaluation results with better model performance and latency show the effectiveness of our approach that combines an efficient approximation and adaptive weight layers.

%\mulh includes weight layers around the approximated similarity, which help adjust the similarity for better model performance, as shown in the paper. This implies an insight that our approach, efficient approximation and learned similarity, is an effective approach to replace cosine similarity in cost volume.

%normalizes the magnitude of the vectors.  Thus, we use 2D layer norm to normalize the vectors along the channel dimension. It ensures that the elements within a pixel vector have a mean of zero and a standard deviation of one. The numerator is the dot product of the vectors. We still use the dot product of the normalized vectors as the similarity. This replacement is not mathematically equivalent, yet the new approach remains effective for similarity measurement.

%Furthermore, the layer norm provides a buffer for the cost volume block, enabling it to accept additional encoded information. The pixel of input activation is normalized so that it can be merged with other encodings. For instance, it can handle positional encoding, which we will introduce in \cref{sec:homo_struct}

%-------------------------------------------------------------------------
\subsection{Multi-head Cost Volume}
\label{subsec:mh_costvolume}
\input{figures/cv_compare}

As another enhancement, we introduce the multi-head attention and dot scale \cite{attention} into our cost volume since separating the input activation in multiple heads enable more perceptions of the network. The approach also leads to a hardware-friendly operator, group-wise dot product followed by a point-wise convolution, as shown in \cref{fig:cv_compare} (b). This replaces the costly cosine similarity depicted in \cref{fig:cv_compare} (a), while providing better model performance. We describe the multi-head cost volume algorithm in a pseudo-code in \cref{alg:MH_cost_volume}.

%Inspired by multi-head attention mechanism \cite{attention}, we introduce the multi-head attention and dot scale into our cost volume since separating the input activation in multiple heads can enable more perceptions of the network. The original cost volume calculates the cosine similarity of pixel vectors of two input activations, and the output is scalar, as shown in \cref{fig:cv_compare_1}. The multi-head cost volume computes dot products to produce numerous scalars, forming new pixel vectors. Subsequently, we apply point-wise convolution to generate an output that matches the size of the original cost volume, as \cref{fig:cv_compare_2} shows. Overall, these operations can be executed just by group-wise Conv, which is highly optimized in modern GPUs. The pseudo-code of multi-head cost volume is shown as \cref{alg:MH_cost_volume}.

\begin{figure}[t]
        \centering
        \includegraphics[width=\linewidth]{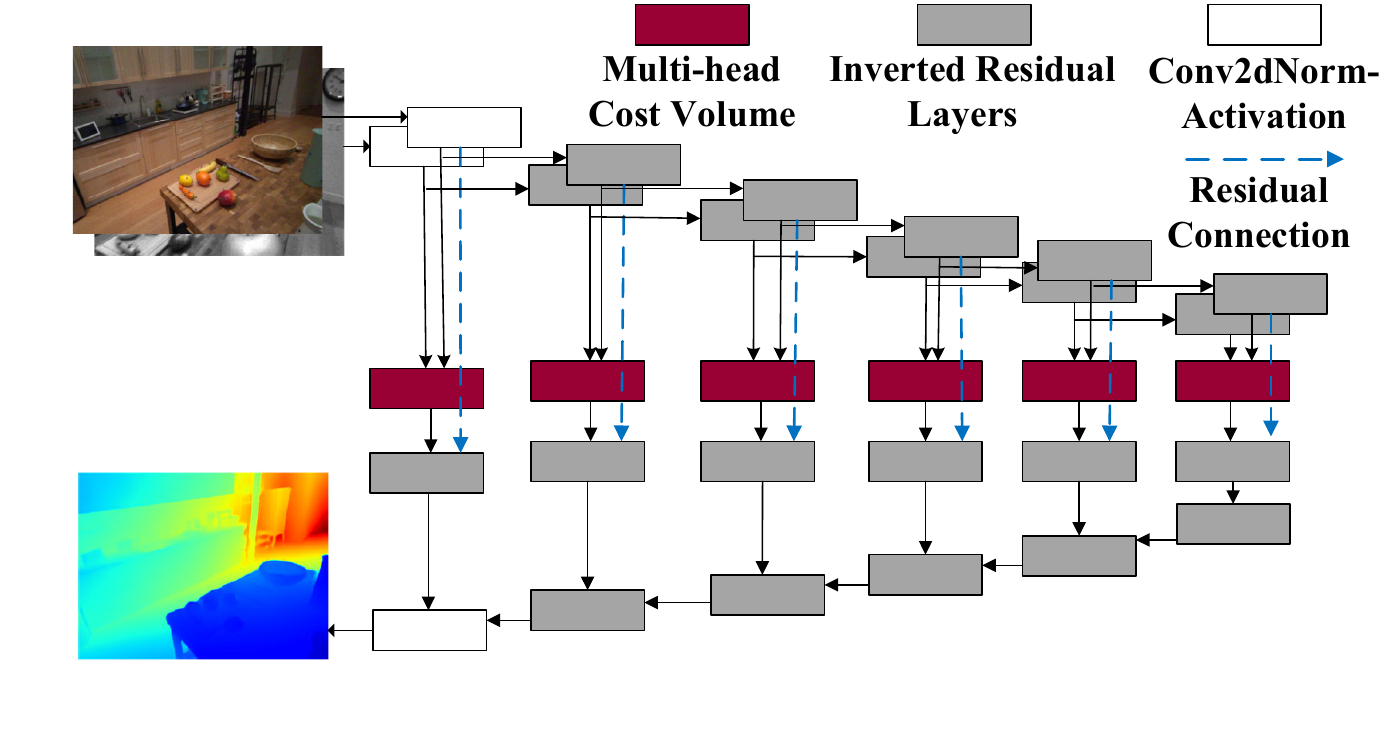}
        \vspace{-8mm}
        \caption{\small \textbf{The structure of \mulh}. The dashed lines indicate that the input activations from the left image are passed to the decoders. The input example is from ADT dataset.}
        \label{fig:mulh}
        \vspace{-4mm}
\end{figure}

\input{algorithms/MH_cost_volume}

The benefits of multi-head cost volume can be summarized with (1) less computational complexity, (2) replacing the computation with highly-optimized group-pointwise Conv, and (3) potential accuracy improvement by providing more perceptions of the input activations.

%-------------------------------------------------------------------------
\subsection{Structure of \mulh}

Applying our approaches discussed in \cref{subsec:apprx_cosine} and \cref{subsec:mh_costvolume} to \argos~\cite{wang2023practical}, we design \mulh, which is illustrated in \cref{fig:mulh}. All settings for the inverted residual blocks \cite{sandler2019mobilenetv2} in \argos are retained to highlight the efficacy of our Multi-head Cost Volume in an ablation study. Note that we apply our approach to Argos since it is the state-of-the-art today; our approach can be applied to any other depth estimation models with cost volume as well. 

Upon this model, \mulh, we add an optimization targeting preprocessing, which we discuss next.

%Replacing the multi-head cost volume blocks with the original cost volume blocks in \cref{fig:mulh.png} results in the structure of \argos described in this paper.

%The structure is shown as \cref{fig:mulh.png}. The encoders are siamese networks composed of inverted residual blocks. The multi-head cost volumes take feature maps from the left and right inputs, and their output retains the same dimensions as the original cost volumes, allowing the decoding setup to remain consistent with that of \argos. The output from the cost volume, along with the encoded left feature map, is passed to the decoding part. The final predicted depth map is aligned to match the left image.

% \insertFigure{mulh.pdf}{\textbf{The structure of \mulh}. The dashed lines indicate that the input activations from the left image are passed to the decoders. The input example is from ADT dataset. \vspace{-5mm}}

%The structure of \mulh is inherited from \argos. All settings for the inverted residual blocks \cite{sandler2019mobilenetv2} in \argos are retained to clearly highlight the differences in the ablation study focused on multi-head cost volume. Replacing the multi-head cost volume blocks with the original cost volume blocks in \cref{fig:mulh.png} results in the structure of \argos described in this paper.

%% file: figures/cv_compare.tex
\begin{figure}[t]
        \centering
        \begin{subfigure}{0.9\linewidth}
            \centering
            \includegraphics[height=0.13\textheight]{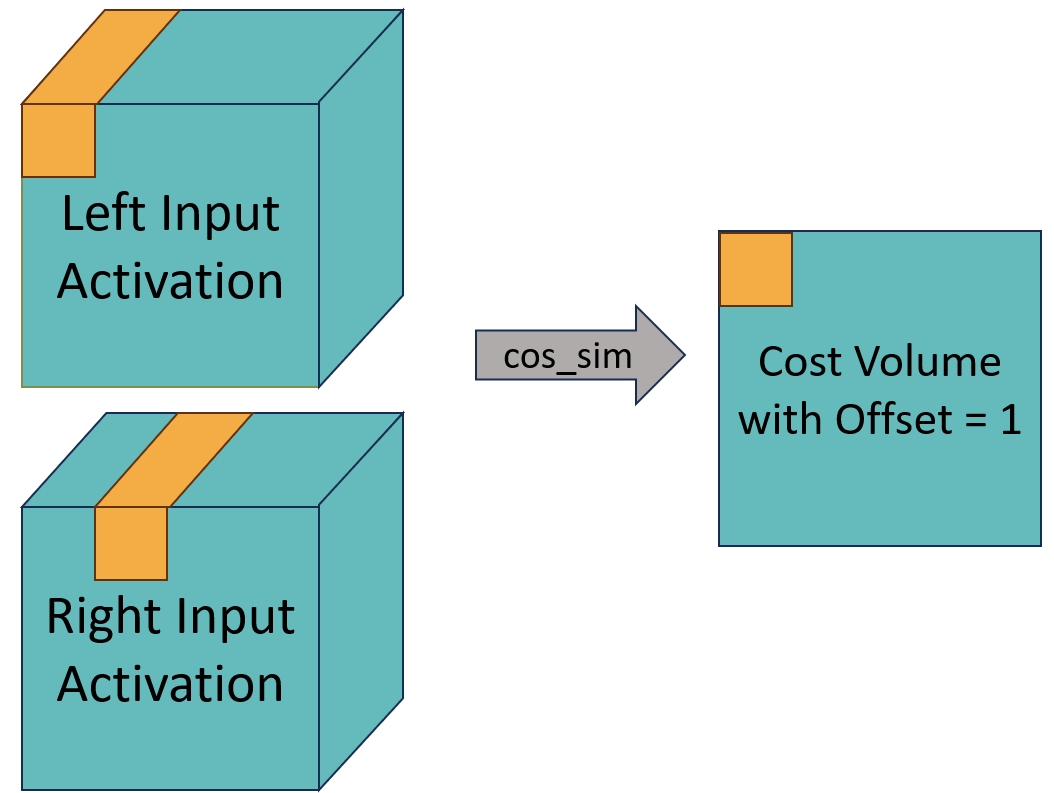}
            \caption{\small \textbf{Diagram of the cost volume with an offset of 1. }
            The orange blocks indicate pixel vectors in the input activations, whose size is $1\times number\_of\_channels$.}
            \label{fig:cv_compare_1}
        \end{subfigure}
        \hfill 

        \begin{subfigure}{0.9\linewidth}
            \centering
            \includegraphics[height=0.13\textheight]{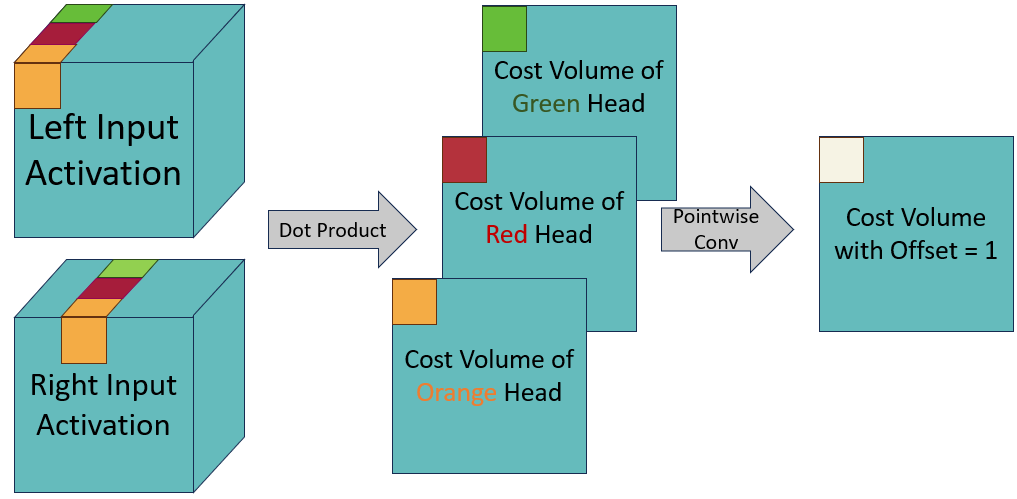}
            \caption{\small \textbf{Diagram of the multi-head cost volume (3 heads in this case) with an offset of 1.} 
            The orange blocks indicate part of pixel vectors, whose size is $1\times \frac{number\_of\_channels}{number\_of\_heads}$. \\ ~}
            \label{fig:cv_compare_2}
        \end{subfigure}
        \vspace{-6mm}
        \caption{\small The comparison between the original cost volume and our approach, \mulhc}
        \label{fig:cv_compare}
        \vspace{-6mm}
    \end{figure}

%% file: algorithms/MH_cost_volume.tex
\begin{algorithm}
\small
\caption{Multi-head Cost Volume}
\label{alg:MH_cost_volume}
\begin{algorithmic}
\Require 
\State $left: [NCHW], ~~ right: [NCHW], ~~ max\_disparity: int,$ 
\State $head\_num: int$ \textcolor{Green}{\Comment{number of heads, can divide C evenly}}
\Ensure 
\State $cost\_volume: [N, max\_disparity, H, W]$
\vspace{2mm} 
\State $s = C/head\_num$ \textcolor{Green}{\Comment{stride of each head}}
\State $left\_norm = \text{LayerNorm}(left)$
\State $right\_norm = \text{LayerNorm}(right)$
\For{$i \in 1:max\_disparity+1$}
    \State $right\_shifted = \text{roll}(right\_norm, i)$ 
    \textcolor{Green}{\Comment{roll: right-shifts the image by \textit{i} pixels, and the right-most \textit{i} columns move to the left}}
    \color{Blue} \For{$h \in 0:head\_num$}
    \State$left\_slice = left\_norm[:, hs:(h+1)s, :, :]$
    \State$right\_slice = right\_shifted[:, hs:(h+1)s, :, :]$
    \State $similarity[:, h, :, :] = \text{dot}(left\_slice, right\_slice)$
    \EndFor
    \State$cost\_volume[:, i, :, :] = \text{pointwise\_conv}(similarity)$
\color{Black} \EndFor
\end{algorithmic}
* In real code, the \textcolor{Blue}{second loop and pointwise\_conv} are replaced by the memory operations and group-pointwise convolution

\end{algorithm}

%% file: sections/04_HomoDepth.tex
\section{HomoDepth Model}
\label{sec:homo}

\begin{figure*}[t]
        \centering
        \includegraphics[width=\textwidth]{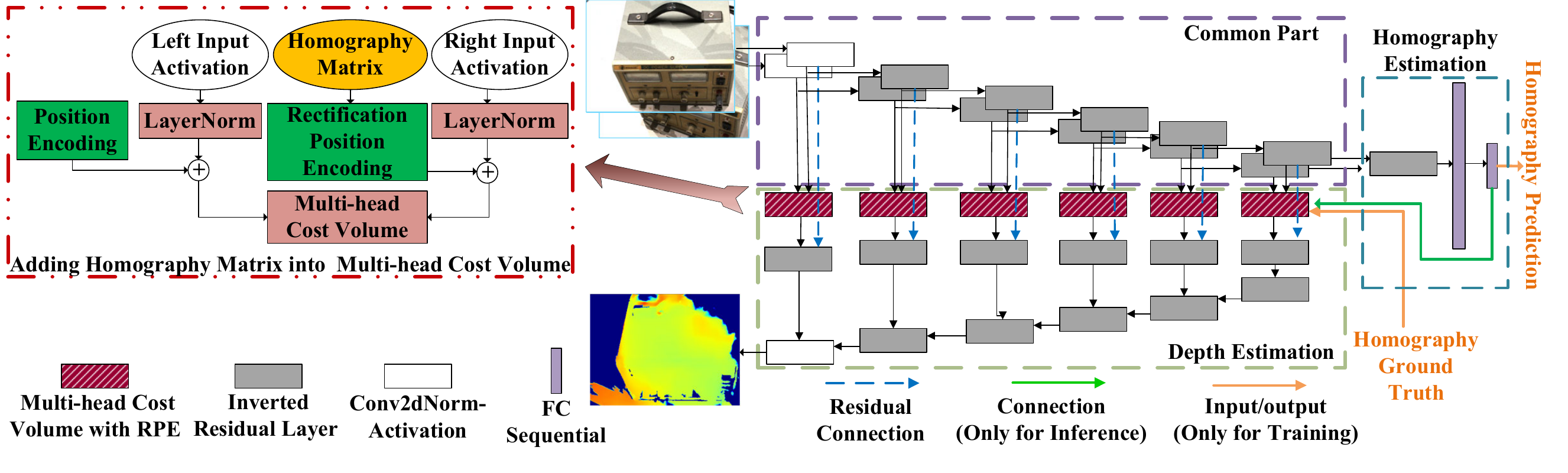}
        \vspace{-10mm}
        \caption{\small \textbf{Structure of \homo.} The input example is from DTU dataset.}
        \label{fig:homo_stru}
        \vspace{-6mm}
\end{figure*}

We discuss how we augment \mulh with an optimization for preprocessing. We first introduce the 3D projection and homographic matrix and discuss how we estimate it using a homography estimation head.

%We discuss how we can augment \mulh to include optimizations for preprocessing, which consists of the introduction of 3D projection and homographic matrix with homography estimation head.

\subsection{3D Projection \& Homographic Matrix}
\label{sec:3dproj_homo}

The main functionality of cost volume is stereo-matching, which evaluates the similarity of left and right inputs with different offsets to determine the possible disparity. Because the algorithm only allows horizontal offsets, it can only match patterns on a common horizontal line. Therefore, preprocessing is required for the input images for stereo cameras without good alignments.
We discuss depth estimation from a broader perspective, focusing on the projection relationship of a world point in the image planes of a stereo camera system. Aligned with that, we pose a fundamental question: \textbf{Given that the image points $q_l$ and $q_r$ are derived from the same world point $Q$, what is the relationship between $q_l$ and $q_r$?}

Suppose the distance from $Q$ to the left and right image planes, intrinsic parameters, and extrinsic parameters are ($d_l, d_r$), ($K_l, K_r$), and ($M_l, M_r$), respectively. Then the the relationship between $q_l$ and $Q$, $q_r$ and $Q$ is as follows:

%Suppose the distance from $Q$ to the left and right image planes are $d_l$ and $d_r$, respectively, intrinsic parameters of left and right cameras are $K_l, K_r$ respectively, and the extrinsic parameters of left and right cameras are $M_l, M_r$ respectively. We derive the relationship between $q_l$ and $Q$, $q_r$ and $Q$,

\begin{equation}
\label{eq:proj_1}
q_l = \frac{1}{d_l}K_lM_lQ  ~~;~~
q_r = \frac{1}{d_r}K_rM_rQ
\end{equation}

Combining them, we obtain

\begin{equation}
\label{eq:proj_2}
q_r = \frac{d_l}{d_r}K_rM_rM_l^{-1}K_l^{-1}q_l=\frac{d_l}{d_r}H_{l\rightarrow r}q_l
\end{equation}

\noindent where $H_{l\rightarrow r}$ is the $3\times 3$ plane homography matrix \cite{textbook}, which  converts $q_l$ to $q_r$. That is, it is easy to represent the positional relation between the content in stereo images if we know the homographic matrix and distance information.

Unfortunately, distance information is unknown in the depth estimation task since it is the eventual goal of the task. However, we can still leverage the following property: ${d_l}/{d_r}\approx1$ when $Q$ is on the central plane of the cameras or far from the cameras. Note that most imaged objects maintain a certain distance from the imaging system, in practice.
%
%But the good thing is ${d_l}/{d_r}$ is the ratio. When $Q$ is on the central plane of the cameras, ${d_l}/{d_r}=1$. Or when $Q$ is far from the camera system, ${d_l}/{d_r}$ is close to 1. In practice, most imaged objects maintain a certain distance from the imaging system. 
%
Thus, homography matrix can approximately represent the positional relationship of stereo image planes, which is helpful for rectification. Homography is widely used in stereo vision applications, and for example, MVSNet \cite{mvsnet} has demonstrated the effectiveness of homography.% In \cref{fig:homo_rect} (b), we can observe homography rectification delivers a desired result. 
%We pick a single cuboid object to highlight the relationships of the images. Because the backgrounds of the images are black and white. We set blue to indicate the margins after transformation. We can observe a distinct feature: the front edge of the cuboid in both images appear parallel, making them suitable for the next steps, such as stereo matching.

MVSNet solves for homography under the assumption that the camera parameters are known but the depth is unknown. Unlike MVSNet, we target more challenging and practical scenarios where both camera extrinsic parameters and depth are unknown. We discuss our methodology to obtain the homography under such conditions next.

%In contrast, our current scenario goes a step further, where both the camera extrinsic parameters and the depth are unknown, and we aim to obtain the homography under these conditions in \cref{sec:homo_est}

%An example of homography rectification is shown as \cref{fig:homo_rect}. We pick a single cuboid object to highlight the relationships of the images. Because the backgrounds of the images are black and white. We set blue to indicate the margins after transformation. We can observe a distinct feature: the front edge of the cuboid in both images appear parallel, making them suitable for the next steps, such as stereo matching.

%-----------------------------------------------------------------
\subsection{Homography Estimation Head}
\label{sec:homo_est}

One challenge from unstable stereo vision systems like AR glasses is that extrinsic parameters keep changing. That is, we can not get the homography matrix directly from \cref{eq:proj_2}. As one solution, DeTone \etal \cite{homography} utilized a convolutional neural network (CNN) for homography estimation. Following the intuition, we also designed a CNN for estimating homography matrix. Unlike the previous work, which designed a stand-alone network dedicated for homography matrix estimation, our homography head shares the encoder with depth estimation, which is trained using a multi-task training technique discussed in~\cref{subsec:multitask_training}.

\subsection{2D Rectification Postional Encoding (RPE)}
\label{subsec:rpe}

Unlike classical epipolar rectification, or the fast online rectification proposed by Argos~\cite{wang2023practical}, homography rectification only needs to transfer one image for alignment. However, one challenge is the lost image information due to margins introduced in the rectified images.

Therefore, to eliminate the needs for homography rectification, we introduce a 2D rectification positional enconding (RPE) methodology that represents the position relationship between stereo images. This approach converts the homography matrix into positional encoding, which involves no information loss unlike  homography rectification. Following \cite{attention}, we define the 2D positional encoding (PE) as:

%Comparing with the classical epipolar rectification, or the fast online rectification proposed by Wang \etal, homography rectification only needs to transfer one image for alignment. But it still generates the margins which lose image information. Thus, we employ the 2D rectification position encoding (RPE) to represent the position relationship between stereo images. It gets rid of transferring the input image. It also converts the homography matrix into positional encoding, making it manageable by CNN.

\vspace{-6mm}

\begin{align}
    \label{eq:pos_enc}
    PE_i(q)&=PE_i(x, y)=
    \begin{cases}
    sin\frac{x}{f^{i/d}}  & if~i = 4k \\
    cos\frac{x}{f^{(i-1)/d}}  & if~i=4k+1 \\
    sin\frac{y}{f^{i/d}}  & if~i=4k+2 \\
    cos\frac{y}{f^{(i-1)/d}}  & if~i=4k+3 \\
    \end{cases}
    \\ i &\in \{4k \leq i \leq 4k+3 | i \in \mathbb{N}, k\in \mathbb{N}, k \in [0, C/4)\} \notag
\end{align}
% \begin{align}
%     \label{eq:pos_enc}
%     PE_i(q)&=PE_i(x, y):=
%     \begin{cases}
%     sin(x~/~ f^\frac{i}{d})  & if~i = 4k \\
%     cos(x~/~f^\frac{i-1}{d})  & if~i=4k+1 \\
%     sin(y~/~f^\frac{i}{d}) & if~i=4k+2 \\
%     cos(y~/~f^\frac{(i-1)}{d})  & if~i=4k+3 \\
%     \end{cases}
%     \\ i &\in \{4k \leq i \leq 4k+3 | i \in N, k\in N, k \in [0, C/4)\} \notag
% \end{align}

\noindent where $q$ is a pixel located at coordinate $(x, y). ~C$ is the number of channels of input activation. $i$ is the channel index. $f$ is the encoding frequency. The value of $f$ depends on the length of the sequence $len$ or the size of the input activation. Generally, it should satisfy $2\pi f > len$, and we default as $f=200$. Here $\mathbb{N}$ means the set of natural numbers.

Note that $(x, y)$ are the coordinates of pixels. And \cref{eq:proj_2} reveals the relationship between $q_l$ and $q_r$. We follow \cref{eq:pos_enc} to get the positional encoding of $q_l$, noted as $PE(q_l)$. Then we apply 2D rectification positional encoding (RPE) to $q_r$ as follows:
\vspace{-6mm}

\begin{equation}
    \label{eq:RPE}
    RPE(q_r):=PE(H_{l \rightarrow r}q_l)
\end{equation}

Because both $PE(q_l)$ and $RPE(q_r)$ use the same coordinates, \textbf{the same location in the real world leads to similar position encoding values in stereo images.} Without RPE, stereo matching solely relies on images semantic similarity. However, RPE incorporates positional similarity, which enhances the similarity across two image points from the same world point utilizing the additional positional information.

%That is, if two image points are projections from the same world point, their similarity increases due to the additional positional information.

\begin{figure}[t]
        \centering
        \includegraphics[width=0.4\textwidth]{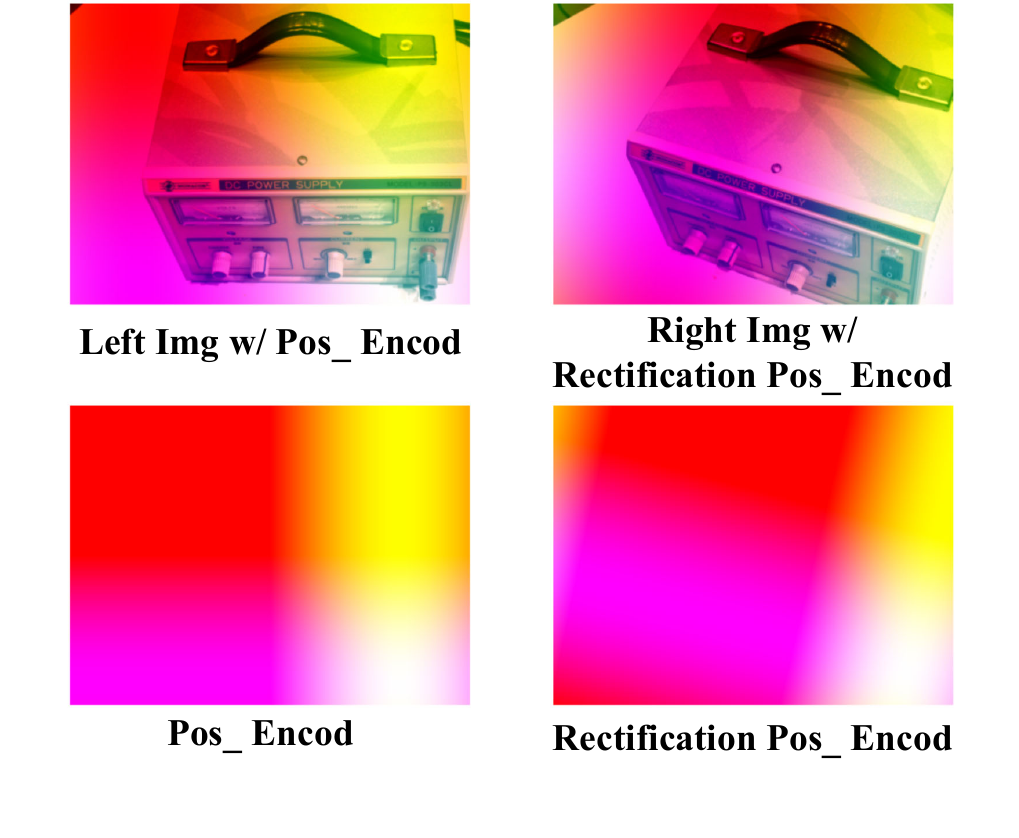}
        \vspace{-9mm}
        \caption{\small \textbf{Illustration of 2D rectification position encoding.} Color patterns indicate the values of positional encodings.}
        \label{fig:pos_enc}
        \vspace{-6mm}
\end{figure}

We present an example of positional encoding in \cref{fig:pos_enc} using colors representing regions with matched encoding values. The left and right images are obtained under a given homography using \cref{eq:pos_enc} and \cref{eq:RPE}, respectively. We observe that the yellow pattern consistently appears on the right buckle of the leather strap, while the front edge of the power supply is always positioned between the red and pink patterns. The results show the efficacy of RPE in capturing alignment information in given stereo images or activations.

\input{tables/dataset_single_column}
\begin{figure*}[]
        \centering
        \includegraphics[width=\textwidth]{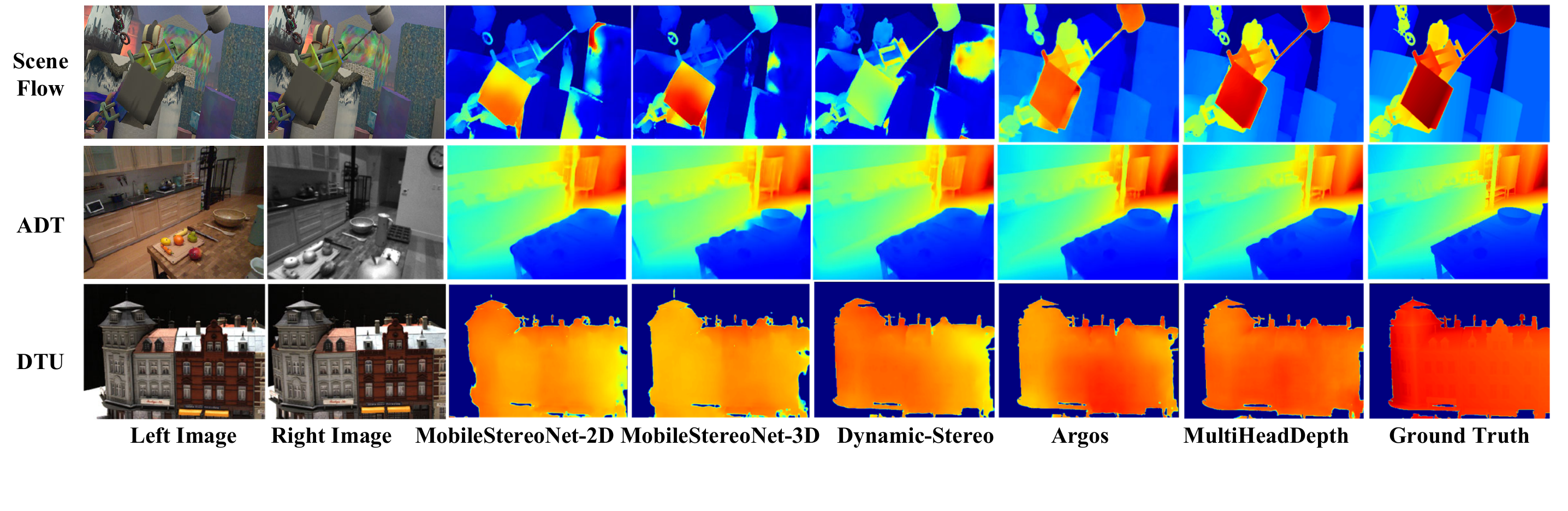}
        \vspace{-15mm}
        \caption{\small A comparison of output qualities across models and datasets}
        \label{fig:res_vis}
        \vspace{-6mm}
\end{figure*}

%-----------------------------------------------------------------
\subsection{Structure of \homo}
\label{sec:homo_struct}

Integrating our approaches discussed in ~\cref{sec:homo_est} and ~\cref{subsec:rpe} to \mulh, we develop \homo, which is illustrated in ~\cref{fig:homo_stru}. One key feature is the multi-task, which generates homography and depth estimation results using two heads connected to the encoder. Also, the RPE is integrated into the multi-head cost volume blocks and estimated homography matrix is used as an input to the multi-head cost volume block, as shown in~\cref{fig:homo_stru}.

%Note that the estimated homography matrix is utilized as inputs of RPE blocks.

%One modification made to the \mulh for the integration is in the multi-head cost volume to accept homography matrix.

%In this section, we are going to introduce the structure of \homo. And the detail of adding homography matrix into the the multi-head cost volume. 

%The structure of \homo is shown as \cref{fig:homo_stru}. Compared with \mulh, the structure just got a few additional blocks. However, with the help of RPE and the multi-task learning method, depth and homography estimation can be finished. As \cref{eq:RPE} shows, the estimated homography matrix is the input of RPE.

In \cref{fig:homo_stru}, we highlight training and inference flows using orange and green arrows, respectively. In the inference phase, the predicted homography matrix is passed to all the multi-head cost volumes, and the RPE is calculated inside them. The range of positional encoding is between 0 and 1, and it is added to the layer norm output. This ensures that the weights of input activations and positional encoding are balanced in the cost volume, allowing similarity to be assessed using both pattern and positional information. In the training phase, the FC block and multi-head cost volume block are disconnected, resulting in two separate inputs and outputs in the model. The inputs include RGB stereo images and their homography matrix, while the outputs are the estimated homography and depth. 

Overall, the model structure is based on three components: a common part (encoder), a depth estimation head, and a homography estimation head, as shown in \cref{fig:homo_stru}. The shared encoder across two heads reduces overall computational costs and enables to perform two tasks with only a minor increase in latency for extra homography estimation head. However, training a multi-task model like ours requires a specialized approach, which we discuss next.

%In \cref{fig:homo_stru}, there are yellow and blue connections that represent the training and inference phases respectively. In the inference phase, the predicted homography matrix is passed to all the multi-head cost volumes, and the RPE is calculated inside it. Since the range of positional encoding is between 0 and 1, it is added to the layer norm output. This ensures that the weights of input activations and positional encoding are balanced in the cost volume, allowing similarity to be assessed using both pattern and positional information.

%During the training phase, the FC block and multi-head cost volume block are disconnected, resulting in two separate inputs and outputs in the model. The inputs include RGB stereo images and their homography matrix, while the outputs are the estimated homography and depth. 

%This structure is divided into three components: a common part, a depth estimation part, and a homography estimation part, as shown in \cref{fig:homo_stru}. The common part reuses the encoding blocks to support both depth and homography estimation, effectively merging the homography estimation and depth estimation networks to conserve computational resources. This design allows \homo to perform multiple tasks with only a minor increase in latency. Training a multi-task model like this requires a specialized approach.

%-----------------------------------------------------------------
\subsection{Multi-task Model Training}
\label{subsec:multitask_training}

To enable multi-task training of \homo, we utilize the homoscedastic uncertainty \cite{MultiTask} to train the model with two loss functions. The first loss function of depth estimation is adopted from \argos \cite{wang2023practical},

\vspace{-5mm}
\begin{align}
    \label{eq:depth_loss}
    L_D(y,\hat{y})&=SL_1(y,\hat{y}) +\sum_{l=0}^4SL_1(\nabla
 y^l, \nabla
\hat{y}^l)
\end{align}

\noindent where $SL_1$ is the SmoothL1Loss \cite{girshick2015fastrcnn}. $\nabla
 y^l$ indicates the gradient of the depth map which is $2 \times 2$ subsampled $l$ times. $y$ is the ground truth, and $\hat{y}$ is the prediction.

The second loss function of homography estimation is, 

\vspace{-2mm}

\begin{equation}
    \label{eq:homo_loss}
    L_H(y,\hat{y})=||weight_w(y)-weight_w(\hat{y})||_F
\end{equation}

where
\vspace{-5mm}

\begin{align}
    \label{eq:homo_weight}
    weight_w(y)=
    \begin{pmatrix}
  w & w & 1 \\
 w & w & 1 \\
  1 & 1 & w
    \end{pmatrix}
    \odot y
\end{align}

In the homography matrix, the first two rows and two columns represent the angular relationship between planes, usually with smaller values. The last column of the first two rows indicates the distance relationship between planes, generally with larger values. The \( weight \) function applies weighting to smaller values by element-wise production, ensuring that all elements of the homography matrix are perceptible in the loss function, which is helpful with better homography matrix estimations. By default, we set $w$ as 50.

Utilizing the loss functions for homography and depth estimation, we formulate combined loss function as:

%In the multi-task learning framework, we combine these into a single total loss function as follows:

\vspace{-2mm}

\begin{equation}
    \label{eq:loss}
    L=\frac{L_H}{2\sigma_H^2}+\frac{L_D}{2\sigma_D^2}+\log \sigma_H \sigma_D
\end{equation}

\noindent where $\sigma_H$ and $\sigma_D$ represent homoscedastic uncertainties of two tasks. They are trainable parameters in \homo.

%% file: tables/dataset_single_column.tex
\begin{table*}[]
\centering
\resizebox{\textwidth}{!}{%
\begin{tabular}{clll}
\hline
\rowcolor[HTML]{C0C0C0} 
{\color[HTML]{000000} \textbf{Dataset}} &
  \multicolumn{1}{c}{\cellcolor[HTML]{C0C0C0}{\color[HTML]{000000} \textbf{SceneFlow}}} &
  \multicolumn{1}{c}{\cellcolor[HTML]{C0C0C0}{\color[HTML]{000000} \textbf{ADT}}} &
  \multicolumn{1}{c}{\cellcolor[HTML]{C0C0C0}{\color[HTML]{000000} \textbf{DTU}}} \\ \hline
\textbf{Scenario} &
  \begin{tabular}[c]{@{}l@{}}Synthetic scenarios (e.g. driving \& flying items)\end{tabular} &
  \begin{tabular}[c]{@{}l@{}}Ego-centric daily life scenarios \end{tabular} &
  \begin{tabular}[c]{@{}l@{}} Robot arm rotating over a single item\end{tabular} \\ \hline
\textbf{\begin{tabular}[c]{@{}c@{}}Images Quality\end{tabular}} &
  \begin{tabular}[c]{@{}l@{}}Synthetic camera-based calibrated RGB images\end{tabular} &
  \begin{tabular}[c]{@{}l@{}}Real camera-based RGB and Grayscale images\end{tabular} &
  \begin{tabular}[c]{@{}l@{}}Real camera-based calibrated  RGB images\end{tabular} \\ \hline
\textbf{\begin{tabular}[c]{@{}c@{}}Stereo Quality\end{tabular}} &
  Rectified &
  Rectified &
  Non-Rectified \\ \hline
\textbf{\begin{tabular}[c]{@{}c@{}}Stereo Baseline\end{tabular}} &
  Static, 1.0 meter &
  Static, 12.8 cm &
  Dynamic, 7.8-14.9cm \\ \hline
\textbf{\begin{tabular}[c]{@{}c@{}}Depth Map\end{tabular}} &
  Obtained by synthetic rendering &
  \begin{tabular}[c]{@{}l@{}}Obtained by twin digital synthetic rendering\end{tabular} &
  \begin{tabular}[c]{@{}l@{}}Rendered from point clouds (need masks for invalid areas)\end{tabular} \\ \hline
\end{tabular}%
}
\vspace{-3mm}
\caption{\small Summary of Datasets}
\vspace{-4mm}
\label{tab:data_sum}
% \vspace{-3mm}
\end{table*}

%% file: sections/05_Evaluation.tex
\input{tables/overall_res}

\section{Evaluation}
\label{sec:evaluation}

We evaluate the effectiveness and efficiency of \mulh and \homo using three datasets on three platforms. We also show the quantization results of the models and discuss the robustness of \homo.

\input{tables/quant_res}

\subsection{Datasets}

We utilize SceneFlow \cite{Sceneflow}, DTU Robot Image Datasets (DTU) \cite{DTU}, and Aria Digital Twin (ADT) \cite{ADT} datasets to evaluate our models and a state-of-the-art stereo depth estimation model for AR glasses, Argos~\cite{wang2023practical}. We summarize the characteristics of the datasets in \cref{tab:data_sum}.
% They are Scene Flow Dataset (SceneFlow) \cite{Sceneflow}, DTU Robot Image Data Sets (DTU) \cite{DTU}, and Aria Digital Twin (ADT) \cite{ADT}. A summary of them is shown as \cref{tab:data_sum}

\noindent \textbf{SceneFlow} SceneFlow is widely used for depth estimation tasks. However, the dataset contains rendered images from synthetic scenarios, which does not fully represent real-world scenarios. Also, its baseline is one meter, which is not aligned with that of AR glasses.

\noindent \textbf{Aria Digital Twin (ADT)} ADT consists of images captured by RGB and gray-scale fisheye cameras mounted on Aria AR glasses~\cite{aria}, prototype AR glasses developed by Meta. Accordingly, the scenarios in the dataset are designed for AR applications. However, the ground truth depth ignores human body in the images. To avoid disturbance from the property, we select 145 out of 236 scenarios that do not involve any human in the scene. %Note that a scenario refers to a video that consists of many frames. 
Due to Aria AR glasses' hardware configuration, each scene in ADT consists of an RGB image from the left and a gray-scale fisheye image from the right. Since both RGB and fisheye images are distorted, we apply calibration before providing images to the models. Also, we replicate the calibrated grayscale images on channel dimension to match the shape of the RGB input tensor. We discuss more details about ADT in \cref{sec:appendix_data}.

%ADT dataset provides one RGB image from the left and one gray-scale fisheye image from the right side for each scene due to the specification of the Aria glasses. Since both RGB and fisheye images are distorted, we apply calibration before feeding images to the models. Also, we replicate the calibrated grayscale images on channel dimension to match the shape of the RGB input tensor. We clarify more details about ADT in \cref{sec:appendix_data}. 

\noindent \textbf{DTU Robot Image Dataset (DTU)} DTU dataset is designed for multi-view stereo (MVS) reconstruction task. The dataset consists of images captured by a camera mounted on a robotic arm performing spherical scans, which moves the robot arm from left to right with a slight downward tilt to the left back upon reaching the edge. We utilize images captured at adjacent positions as stereo image pairs for training. Compared to other MVS datasets, the DTU dataset is closer to the AR glasses scenarios since (1) its scanning trajectory is close to the horizontal line, which can mimic the misalignment of stereo images. (2)  The scanning intervals between adjacent images (i.e., baselines), are $\approx$10 cm, which is similar as the baseline of AR glasses~\cite{aria}.%, making it more suitable for simulating the vision system in AR applications.

%\begin{itemize}
%    \item Its scanning trajectory is close to the horizontal line, which can mimic the misalignment of stereo images.
%    \item The scanning intervals, or the baselines between adjacent images, are around 10 cm. This is similar to the baseline of cameras on AR glasses, making it more suitable for simulating the vision system in AR applications.
%\end{itemize}

We train our model for the three datasets individually and evaluate the performance. Since the DTU contains the extrinsic parameters of each frame, we leverage them in calculating the homography matrix over each stereo image inputs taken from two adjacent frames. We utilize the calculated homography matrix as the ground truth for the homography estimation in \homo. In the accuracy evaluation, we also evaluate our model on Middlebury 2014 \cite{middlebury}.

\begin{figure}[t]
        \centering
        \includegraphics[width=0.9\linewidth]{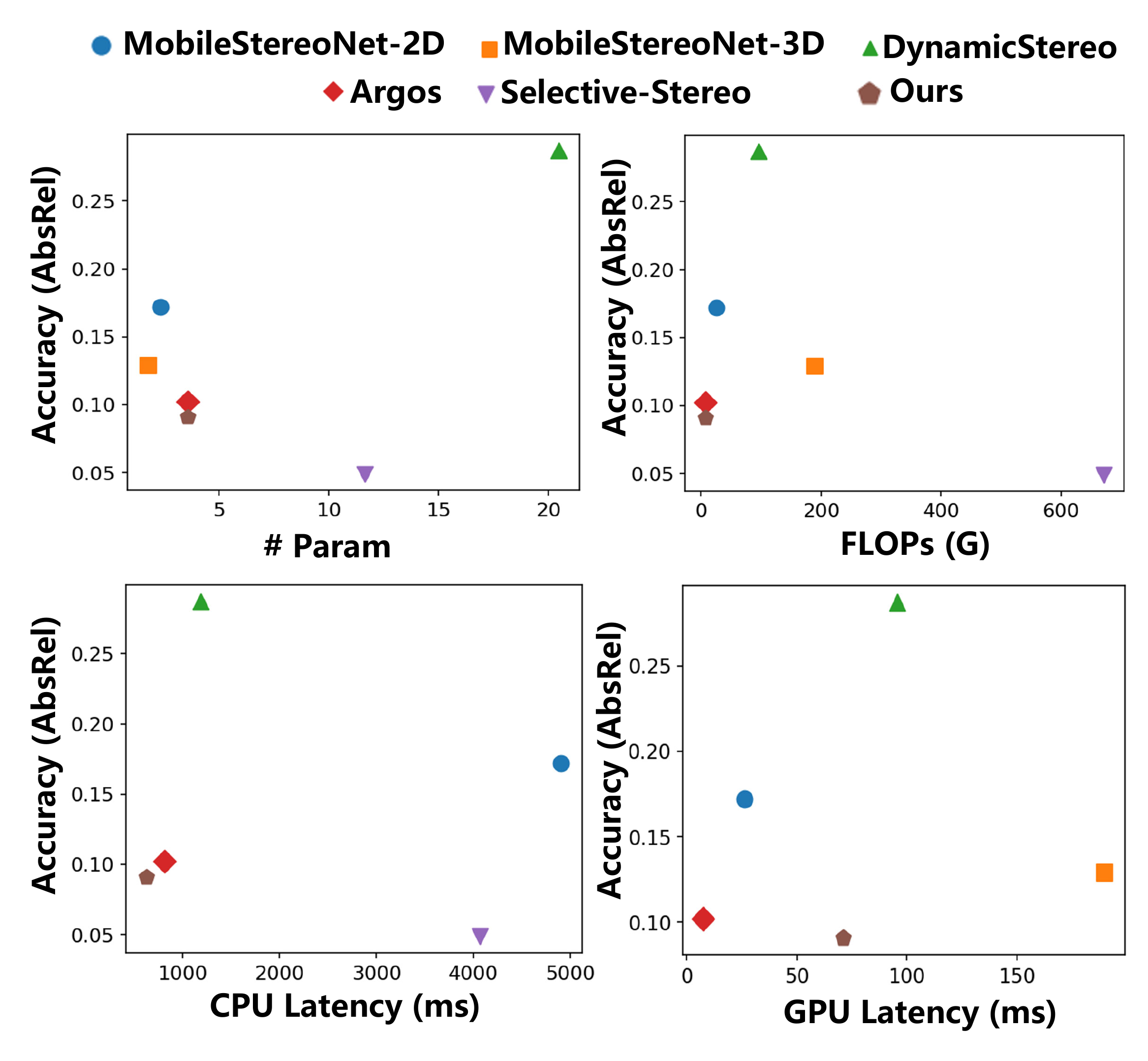}
        \vspace{-6mm}
        \caption{\small \textbf{The overall comparisons of models regarding \# parameters, FLOPs, and latencies versus accuracy.} For all coordinates, the closer to the bottom left, the better model performance. Here CPU indicates Core\texttrademark ~i7-12700H and GPU indicates Nvidia\textregistered ~GeForce RTX 3070 Ti laptop GPU. }
        \label{fig:scatter}
        \vspace{-6mm}
\end{figure}

\input{tables/combine_ablation_edge_runtime}

%--------------------------------------------------------------------

\subsection{Implementation details}
We implement \mulh and \homo in Pytorch \cite{pytorch} and train on Nvidia RTX 4090 24GB GPU. %We run all the models involved in this section to obtain the test results.
We resize all input images in all experiments to $288\times 384\times 3$. We use Adam optimizer without schedulers. In the early stages of model training, the base learning rate (LR) is \textit{1e-4}, and we select the best epoch. Afterward, we fine-tune the model with an LR of \textit{4e-4} and select the optimal weights. We set the batch size as 10, based on the GPU memory.

\subsection{Evaluation Metrics and Platform}

We adopt AbsRel, D1, and RMSE as the evaluation metrics to measure model performance. All the metrics are "Lower is Better" metrics, where smaller values represent higher model performance. We provide the detailed definition of each metric in \cref{sec:appendix_metric}.

Since our models' eventual target is AR glasses, and there is no AR glasses with open API to the best of our knowledge, we use three mobile platforms for latency evaluation as proxy. For the main model performance evaluation, we use a laptop with Intel\textregistered ~Core\texttrademark ~i7-12700H CPU and Nvidia\textregistered ~GeForce RTX 3070 Ti laptop GPU. We also measure the inference latency of our models on Nvidia\textregistered ~Jetson Orin Nano\texttrademark~ Developer Kit \cite{nvidiaorinnano} and a smartphone equipped with Snapdragon 8+ Gen 1 \cite{snape8gen1}.

\subsection{\mulh Performance}

We first evaluate the model performance on calibrated inputs. We compare the accuracy of \mulh against SOTA stereo depth estimation models listed in~\cref{tab:overall}. The overall results plotted in \cref{fig:scatter} demonstrate that \argos and \mulh are superior solutions for evaluated scenarios designed to model AR glasses. Comparing \mulh against \argos, \mulh provides 11.8-30.3\% improvements in accuracy and 22.9-25.2\% reduction in latency. Compared to Selective-Stereo, \mulh achieves competitive accuracy while using only 1.2\% of the FLOPs required by Selective-Stereo. The results show the effectiveness of our approach to optimize the cost volume blocks.

\noindent
\textbf{Impact of Quantization}
%The performance reported for \argos in \cite{wang2023practical} is based on the quantized model. 
As quantization is commonly adopted optimization like Argos~\cite{wang2023practical}, we also evaluate quantized models and present the results in \cref{tab:quant_res}. In this study, we use Aimet \cite{AIMET} from Qualcomm to perform a post-training quantization. Although the overall accuracy decreases slightly by 3.1-14.7\%, the latency is considerably reduced by 7.7-43.4\% after quantization. \mulh remains superior in all metrics after quantization.

%--------------------------------------------------------------------
\subsection{\homo Performance}

%Because the Sceneflow and ADT datasets consist of rectified stereo image inputs, it is meaningless to evaluate their accuracy on \homo. A detailed analysis will be provided in the next section.

Since \homo optimizes preprocessing, we focus on the DTU scenarios that require preprocessing (rectification) for evaluating \homo. % and evaluate models with preprocessing and \homo on such scenarios.
%We evaluate the efficacy of RPE in the model through ablation studies of the end-to-end model (preprocessing + depth estimation). 
Utilizing the DTU dataset, we create a custom dataset, DTU\_df, by scaling and cropping DTU images to simulate varying focal lengths ("df": dynamic focal length). Thus, the DTU dataset represents a stereo system with a fixed focal length but variable relative position, while DTU\_df represents a system with both variable focal length and relative position. Furthermore, we apply perspective transformation on Sceneflow (Sceneflow\_persp) to simulate the misalignment of cameras.

In \cref{tab:ablation}, we evaluate four models with and without preprocessing. Although \mulh requires less inference latency than \homo, it still relies on preprocessing to achieve better accuracy, as shown in \cref{tab:overall} and \cref{tab:ablation}. In contrast, \homo does not require preprocessing. When the stereo cameras are misaligned, \homo becomes a better solution, delivering higher accuracy with 30\% less end-to-end latency.

%--------------------------------------------------------------------
\subsection{Latency on Edge Devices}

% In practice, although AR glasses are equipped with an SoC, they typically rely on cloud or local devices for large-scale computational tasks. This is due to concerns such as computational resource limitation, heat generation, and energy consumption. Compared to cloud computing, local computation offers advantages such as faster data transfer speeds and better user privacy. Local devices include the user's smartphone and specialized wireless compute puck. Therefore, it is essential to evaluate the model latencies on edge devices. 
    
We report the latencies on two mobile/edge scale devices, Nvidia\textregistered ~Jetson Orin Nano\texttrademark~ Developer Kit and Snapdragon 8+ Gen 1 Mobile Platform in \cref{tab:edge_res}. On the Orin Nano, the models are executed in eager mode. On the Snapdragon, the models are compiled using SNPE \cite{qualcomm2024snpe} and executed via ADB commands. The results may vary due to different deployment methods. We observe our models overall achieve better latency compared to Argos. Note that Argos results do not include preprocessing delay, while \homo can accept unrectified data without preprocessing.

%% file: tables/overall_res.tex
\begin{table*}[]
\centering
\resizebox{\textwidth}{!}{%
\begin{tabular}{crrr|rrr|rrr|rrr|rrr|rrr}
\hline
                                         & \multicolumn{3}{c|}{\textbf{\begin{tabular}[c]{@{}c@{}}MobileStereoNet-2D\\ (WACV 2022 \cite{MobileStereoNet})\end{tabular}}} & \multicolumn{3}{c|}{\textbf{\begin{tabular}[c]{@{}c@{}}MobileStereoNet-3D\\ (WACV 2022 \cite{MobileStereoNet})\end{tabular}}} & \multicolumn{3}{c|}{\textbf{\begin{tabular}[c]{@{}c@{}}Dynamic-Stereo \\ (CVPR 2023 \cite{DynamicStereo})\end{tabular}}} & \multicolumn{3}{c|}{\textbf{\begin{tabular}[c]{@{}c@{}}Argos\\ (CVPR 2023 \cite{wang2023practical})\end{tabular}}} & \multicolumn{3}{c|}{\textbf{\begin{tabular}[c]{@{}c@{}}Selective-Stereo\\ (CVPR 2024 \cite{wang2024selective})\end{tabular}}} & \multicolumn{3}{c}{\textbf{\begin{tabular}[c]{@{}c@{}}MultiheadDepth\\ (Ours)\end{tabular}}}              \\ \cline{2-19} 
                                         & \multicolumn{1}{c}{\textbf{AbsRel}}             & \multicolumn{1}{c}{\textbf{D1}}             & \multicolumn{1}{c|}{\textbf{RMSE}}             & \multicolumn{1}{c}{\textbf{AbsRel}}             & \multicolumn{1}{c}{\textbf{D1}}             & \multicolumn{1}{c|}{\textbf{RMSE}}             & \multicolumn{1}{c}{\textbf{AbsRel}}            & \multicolumn{1}{c}{\textbf{D1}}           & \multicolumn{1}{c|}{\textbf{RMSE}}           & \multicolumn{1}{c}{\textbf{AbsRel}}          & \multicolumn{1}{c}{\textbf{D1}}         & \multicolumn{1}{c|}{\textbf{RMSE}}         & \multicolumn{1}{c}{\textbf{AbsRel}}             & \multicolumn{1}{c}{\textbf{D1}}             & \multicolumn{1}{c|}{\textbf{RMSE}}            & \multicolumn{1}{c}{\textbf{AbsRel}} & \multicolumn{1}{c}{\textbf{D1}} & \multicolumn{1}{c}{\textbf{RMSE}} \\ \hline
\multicolumn{1}{c|}{\textbf{Sceneflow}}  & 0.172                                           & 0.71                                        & 8.2                                            & 0.129                                           & 0.42                                        & 5.9                                            & 0.287                                          & 0.89                                      & 68.8                                         & 0.102                                        & 0.23                                    & 5.3                                        & \textbf{0.053}                                  & \textbf{0.07}                               & 21.99                                         & 0.091                               & 0.35                            & \textbf{4.3}                      \\ \hline
\multicolumn{1}{c|}{\textbf{Middlebury}} & 0.180                                           & 0.43                                        & 65.6                                           & 0.139                                           & 0.32                                        & 25.7                                           & 0.151                                          & 0.38                                      & 20.15                                        & 0.107                                        & 0.016                                   & 27.54                                      & 0.169                                           & 0.41                                        & \textbf{12.90}                                & \textbf{0.094}                      & \textbf{0.01}                   & 23.6                              \\ \hline
\multicolumn{1}{c|}{\textbf{ADT}}        & 0.199                                           & \textbf{0.36}                               & 611.9                                          & 0.135                                           & 0.52                                        & 563.3                                          & 0.176                                          & 0.55                                      & 546.1                                        & 0.133                                        & 0.34                                    & 320.1                                      & \textbf{0.082}                                  & 0.25                                        & 277.3                                         & 0.094                               & \textbf{0.25}                   & \textbf{273.6}                    \\ \hline
\multicolumn{1}{c|}{\textbf{DTU}}        & 0.147                                           & \textbf{0.38}                               & 315.9                                          & 0.148                                           & 0.40                                        & 224.8                                          & 0.339                                          & 0.88                                      & 628.5                                        & 0.122                                        & 0.61                                    & 536.9                                      & 0.128                                           & 0.49                                        & 229.7                                         & \textbf{0.101}                      & 0.42                            & \textbf{216.1}                    \\ \hline
\end{tabular}
}
\vspace{-3mm}
\caption{\small \textbf{The overall accuracy of the models.} The datasets are used as inputs without any preprocessing. As indicated in \cref{tab:data_sum}, SceneFlow, Middleburry, and ADT datasets consist of rectified stereo image inputs, while DTU contains unrectified stereo image inputs. For Middleburry, following the practice in \cite{wang2023practical}, models are trained with Sceneflow and tested on Middlebury 2014.}
\label{tab:overall}
\vspace{-5mm}
\end{table*}

%% file: tables/quant_res.tex
\begin{table}[]
\centering
\resizebox{\linewidth}{!}{%
\begin{tabular}{lcrr|cr}
\hline
\multirow{2}{*}{} & \multicolumn{3}{c|}{\textbf{AbsRel of Datasets}} & \multicolumn{2}{c}{\textbf{Latency (ms)}} \\ \cline{2-6} 
 &
  \multicolumn{1}{c}{SceneFlow} &
  \multicolumn{1}{c}{ADT} &
  \multicolumn{1}{r|}{DTU} &
  \multicolumn{1}{c|}{CPU} &
  \multicolumn{1}{c}{GPU} \\ \hline
\multicolumn{1}{c|}{\begin{tabular}[c]{@{}c@{}}\textbf{Argos}    (CVPR 2023)\end{tabular}} &
  \multicolumn{1}{r}{0.109} &
  \multicolumn{1}{r}{0.146} &
  0.140 &
  \multicolumn{1}{r|}{748.5} &
  54.4 \\ \hline
\multicolumn{1}{c|}{\begin{tabular}[c]{@{}c@{}}\textbf{MultiheadDepth} (Ours)\end{tabular}} &
  \multicolumn{1}{r}{\textbf{0.098}} &
  \multicolumn{1}{r}{\textbf{0.097}} &
  \textbf{0.112} &
  \multicolumn{1}{r|}{\textbf{598.9}} &
  \textbf{45.3} \\ \hline
\end{tabular}%
}
\vspace{-3mm}
\caption{\small \textbf{The performance of quantized models (INT8)}. The latencies are measured on a laptop with Intel\textregistered ~Core\texttrademark ~i7-12700H CPU and Nvidia\textregistered ~GeForce RTX 3070 Ti laptop GPU.}
\label{tab:quant_res}
\vspace{-6mm}
\end{table}

%% file: tables/combine_ablation_edge_runtime.tex
\begin{table*}
    \centering
\begin{minipage}{0.66\textwidth}
\centering
\resizebox{\textwidth}{!}{%
\begin{tabular}{cccrr|crr|crr|crr}
\hline
\multicolumn{2}{l}{\multirow{2}{*}{}} & \multicolumn{3}{c|}{\textbf{Argos}} & \multicolumn{3}{c|}{\textbf{PreP+Argos}} & \multicolumn{3}{c|}{\textbf{PreP+MultiHeadDepth}} & \multicolumn{3}{c}{\textbf{HomoDepth}} \\ \cline{3-14} 
\multicolumn{2}{l}{} & \textbf{AbsRel} & \multicolumn{1}{c}{\textbf{D1}} & \multicolumn{1}{c|}{\textbf{RMSE}} & \textbf{AbsRel} & \multicolumn{1}{c}{\textbf{D1}} & \multicolumn{1}{c|}{\textbf{RMSE}} & \textbf{AbsRel} & \multicolumn{1}{c}{\textbf{D1}} & \multicolumn{1}{c|}{\textbf{RMSE}} & \textbf{AbsRel} & \multicolumn{1}{c}{\textbf{D1}} & \multicolumn{1}{c}{\textbf{RMSE}} \\ \hline
\multicolumn{1}{c|}{\multirow{3}{*}{\textbf{Dataset}}} & \multicolumn{1}{c|}{\textbf{DTU}} & \multicolumn{1}{r}{0.122} & 0.61 & 536.91 & \multicolumn{1}{r}{0.109} & 0.54 & 210.8 & \multicolumn{1}{r}{0.099} & \textbf{0.38} & \textbf{204.4} & \multicolumn{1}{r}{\textbf{0.098}} & \textbf{0.38} & 208.8 \\ \cline{2-14} 
\multicolumn{1}{c|}{} & \multicolumn{1}{c|}{\textbf{DTU\_df}} & \multicolumn{1}{r}{0.183} & 0.64 & 599.32 & \multicolumn{1}{r}{0.115} & 0.55 & 223.1 & \multicolumn{1}{r}{0.107} & 0.48 & 287.8 & \multicolumn{1}{r}{\textbf{0.087}} & \textbf{0.32} & \textbf{184.9} \\ \cline{2-14} 
\multicolumn{1}{c|}{} & \multicolumn{1}{l|}{\textbf{Sceneflow\_persp}} & \multicolumn{1}{r}{0.232} & 0.71 & 597.33 & \multicolumn{1}{r}{0.121} & 0.43 & 209.9 & \multicolumn{1}{r}{0.114} & 0.42 & 215.3 & \multicolumn{1}{r}{\textbf{0.097}} & \textbf{0.40} & \textbf{198.4} \\ \hline
\multicolumn{1}{c|}{\multirow{2}{*}{\textbf{\begin{tabular}[c]{@{}c@{}}Latency\\ (ms)\end{tabular}}}} & \multicolumn{1}{c|}{\textbf{CPU}} & \multicolumn{3}{c|}{811.0} & \multicolumn{3}{c|}{1068.3} & \multicolumn{3}{c|}{884.4} & \multicolumn{3}{c}{\textbf{761.2}} \\ \cline{2-14} 
\multicolumn{1}{c|}{} & \multicolumn{1}{c|}{\textbf{GPU}} & \multicolumn{3}{c|}{109.0} & \multicolumn{3}{c|}{312.5} & \multicolumn{3}{c|}{312.6} & \multicolumn{3}{c}{\textbf{84.5}} \\ \hline
\end{tabular}
}
\vspace{-3mm}
\caption{\small \textbf{Ablation study comparing models with or without rectification preprocessing (PreP).} DTU dataset represents a stereo system with a fixed focal length but variable relative position, while DTU\_df represents a system with both variable focal length and position. Sceneflow\_persp is the simulation of glasses bending by perspective transformation.}
\label{tab:ablation}
\end{minipage}
	%\qquad
	\hfill
\begin{minipage}{0.32\textwidth}
\centering
\resizebox{\textwidth}{!}{%
\begin{tabular}{crr|rr}

\hline
\multicolumn{1}{l}{} & \multicolumn{2}{c|}{\textbf{Orin Nano}} & \multicolumn{2}{c}{\textbf{Snapdragon}} \\ \cline{2-5} 
\multicolumn{1}{l}{} & \multicolumn{1}{c}{CPU} & \multicolumn{1}{c|}{GPU} & \multicolumn{1}{c}{CPU} & \multicolumn{1}{c}{GPU} \\ \hline
\multicolumn{1}{c|}{\textbf{Argos} (CVPR 2023)} & \multicolumn{1}{r}{8774} & 209 & \multicolumn{1}{r}{1424} & 914 \\ \hline
\multicolumn{1}{c|}{\textbf{MulHeadDepth} (Ours)} & \multicolumn{1}{r}{\textbf{6183}} & 216 & \multicolumn{1}{r}{\textbf{1156}} & \textbf{617} \\ \hline
\multicolumn{1}{c|}{\textbf{HomoDepth} (Ours)} & \multicolumn{1}{r}{6611} & \textbf{203} & \multicolumn{1}{r}{1512} & 893 \\ \hline
\end{tabular}
}
\vspace{-3mm}
\caption{\small \textbf{Latency of models in millisecond} on NVIDIA\textregistered ~Jetson Orin Nano\texttrademark~ Developer Kit and Snapdragon 8+ Gen 1 Platform.}
\label{tab:edge_res}
\end{minipage}
\vspace{-5mm}
\end{table*}

%% file: sections/06_Conclusion.tex
\section{Conclusion}
\label{sec:conclusion}

Achieving both high model performance and low latency is a key toward depth estimation on AR glasses. In this work, we identify cost volume and preprocessing as the major optimization targets for low latency. %By analyzing hardware-friendly and -unfriendly operations in existing cost volume,
For cost volume, we develop a hardware-friendly approximation of cosine similarity and employ a group-pointwise convolution. For preprocessing, we introduce rectification positional encoding (RPE) and leverage homography matrix estimated by a head attached to the common encoder in \homo, which significantly reduces the computational complexity.

Our approaches are complementary to any stereo depth estimation models with online rectification and/or cost volume. In addition to the broad applicability, our evaluation results show the effectiveness of our approaches in model performance as well as latency. Therefore, we believe our work is making a meaningful step forward in the field of stereo depth estimation systems for AR.

\section{Acknowledgment}

We thank Dilin Wang, Wei Ye, and Lianghzen Lai for their insightful feedback, Jialiang Wang for the guidance in modeling Argos, Meta for providing Aria AR glasses, and Rachid Karami for the assistance with the latency analysis.

%% file: sections/10_Appendix.tex
\clearpage

\appendix
    
\section{Computational Complexity of Multi-head Cost Volume}
\label{subsec:compute_complexity_mhcv}

Suppose the dimension of an input activation is $CHW$, and the max disparity is $d$. The number of multiply-accumulate operations (\#MAC) of the original cost volume is $3CHWd$ (please refer to~\cref{alg:MH_cost_volume}). The layer norm along channel dimension is defined as \cref{eq:layernorm}, whose \#MAC is $3HWd$. Replacing the cosine similarity with the dot product, and adding the layer norm before the loop reduce the \#MAC. The \#MAC of multi-head cost volume is $2\times 3CHW+CHWd<3CHWd$. (The definition of parameters in \cref{eq:layernorm} follow \cite{layernorm}.)
\begin{equation}
    y=\frac{x-E[x]}{\sqrt{Var[x]+\epsilon}}\gamma+\beta
    \label{eq:layernorm}
\end{equation}

%-------------------------------------------------------------------
% \section{Approximation about the Cosine Similarity}
% \label{subsec:appendix_apprx_cosine}

%-------------------------------------------------------------------
\section{SIFT v.s. Conv Network}
\label{sec:appendix_sift}

As \cite{vinukonda2011study} analyses, the computational complexity of SIFT for an $N \times N$ image with $n\times n$ tiles is
\begin{equation}
\label{eq:sift_complex}
    \Theta \left( \frac{(n + x)^2}{p_i \, \Gamma_0} + \alpha \beta N^2 x^2 \Gamma_1 + \frac{(\alpha \beta + \gamma) n^2 \log x}{p_o \, \Gamma_2} \right)
\end{equation}

\noindent where x is the neighborhood of tiles. $N\gg n>x$ in most cases, we can simplify the complexity as $O(N^2)$.

For convolutional networks, the computational complexity for an $N \times N$ input activation with $C$ channels in input and $C'$ channels in output is,

\begin{equation}
\label{eq:conv_complex}
    O(N^2CC'k^2)
\end{equation}

\noindent where $k$ is the convolution kernel size.  $N\gg C>k$ in most cases, we can also simplify the complexity as $O(N^2)$.

Based on the above analysis, we can conclude that: \textbf{Supervised learning convolutional neural networks capable of the same task will not perform worse efficiency for all computer vision algorithms requiring key point matching.} As the same, CNN is as efficient as, or even outperforms, classic algorithms in homography estimation tasks.

In practice, many optimizations for CNNs have been proposed, and CNN computations are more hardware-friendly. In contrast, \cite{vinukonda2011study} has been proven that without improvements in the input bandwidth, the power of multicore processing cannot be used efficiently for SIFT. Therefore, CNNs are generally a more efficient approach. Based on the report in \cite{wang2023practical} and our experiments, keypoint matching takes around 300ms on both smartphones and laptops. There is no significant speed up from smartphone to laptop, showing the limitations of keypoint matching.

%-------------------------------------------------------------------
% \section{Epipolar Rectification v.s. Homography Rectification}
% \cerf{} is the examples about the epipolar rectification and homography rectification to show their differences
% \input{figures/homo_rect}

%-------------------------------------------------------------------
\section{Dataset Setting}
\label{sec:appendix_data}

\paragraph{DTU setting:} Based on previous implementations and common practices \cite{mvsnet}, we selected the evaluation set as scans \{1, 4, 9, 10, 11, 12, 13, 15, 23, 24, 29, 32, 33, 34, 48, 49, 62, 75, 77, 110, 114, 118\}, validation set: scans \{3, 5, 17, 21, 28, 35, 37, 38, 40, 43, 56, 59, 66, 67, 82, 86, 106, 117\}, and the rest is training set. 

Additionally, our network takes two images as input, but the depth map is aligned with the left image. This means that the left and right inputs cannot be interchanged. We need to match the stereo images to ensure that the relative position of the left input is indeed on the left side of the right input. The image match list is shown as \cref{tab:dtu_pair}.

\paragraph{ADT setting:} As mentioned, ATD ignores users' bodies in the images when rendering ground truth depth maps which causes inconsistencies between the input images and the predicted results. We selected subsets of the scene where no other users were present. The selected subset can be obtained by this query link: \url{https://explorer.projectaria.com/adt?q=%22is_multi_person+%3D%3D+false%22}

\input{tables/dtu_pair}

%-------------------------------------------------------------------
\section{Metrics Definition}
\label{sec:appendix_metric}
Here are the definitions of the metrics we used for evaluation:
\begin{align}
    AbsRel &= \frac{1}{N} \sum_{i=1}^{N} \frac{\hat y_i-y_i}{y_i} \\
    D1 &= \frac{1}{N} \sum_{i=1}^{N} \mathbf{1} (\frac{|\hat{y}_i - y_i|}{y_i} \leq 5 \% ) \\
    RMSE &= \sqrt{\frac{1}{N} \sum_{i=1}^N (\hat{y}_i-y_i)^2}
\end{align}

\noindent where $N$ is the total number of pixels of the depth maps, $y_i$ is the value of pixels in ground truth map, and $\hat{y_i}$ is the value of pixels in prediction.

\section{Robustness Analysis}
\homo is a muti-task learning structure, and the depth estimation depends on the predicted homography matrix. Thus, the accuracy of homography estimation affects depth estimation. In this section, we address two questions: (1) How sensitive is the depth estimation to the homography estimation? (2) How stable is the homography estimation?

%\begin{itemize}
%    \item How much the homography estimation will affect the depth estimation?
%    \item How stable is the homography estimation?
%\end{itemize}

\paragraph{Homography Estimation Errors.} We analyze the error of \homo during homography estimation. In practice, the error variance is as small as 0.003, as shown in \cref{fig:homo_stable_2}. This indicates that the homography estimation of \homo is highly accurate.

\paragraph{Sensitivity Study.} In \homo, we inject noise \( n \sim N(0, \sigma) \), where \( \sigma \in [0, 2] \), to the elements of estimated homography matrix before it is passed to the multi-head cost volume blocks. Then, we investigate the final depth estimation errors. The instability is quantified by examining the noise variance and corresponding changes in the smooth loss function \cref{eq:depth_loss} corresponding to depth estimation. As shown in \cref{fig:homo_stable_1}, the standard deviation trends indicate that depth estimation remains stable when $\sigma < 0.5$. The linear fitting demonstrates that the loss values increase as the noise variance grows.

%. Before the predicted homography matrix is passed to the cost volumes, we add noise \( n \sim N(0, \sigma) \), where \( \sigma \in [0, 2] \), to the elements of the matrix. We then evaluate how the errors in the homography matrix prediction impact the final depth estimation. The instability is quantified by examining the noise variance and corresponding changes in the smooth loss function \cref{eq:depth_loss} corresponding to depth estimation. As shown in \cref{fig:homo_stable_1}, the standard deviation trends indicate that depth estimation remains stable when $\sigma < 0.5$. The linear fitting demonstrates that the loss values increase as the noise variance grows.

\input{figures/homo_stable}

\section{Application Scenarios}
For one-shot scenarios, we recommend using \homo. For continuous frames scenarios, we assume that the relative positions of the cameras on AR glasses remain stable over a short period. Therefore, we recommend first running \homo to obtain depth estimation while simultaneously deriving the homography between the two cameras. For subsequent stereo inputs, rectification can be quickly applied with homography, and \mulh can be utilized to achieve higher efficiency.

%% file: tables/dtu_pair.tex
\begin{table}[]
\resizebox{0.45\textwidth}{!}{%
\begin{tabular}{
>{\columncolor[HTML]{C0C0C0}}c |cccccccccccc}
\hline
L & 1  & 2  & 3  & 4  & 5  & 7  & 8  & 9  & 10 & 11 & 12 & 12 \\
R & 2  & 3  & 4  & 5  & 6  & 6  & 7  & 8  & 9  & 10 & 11 & 13 \\ \hline
L & 13 & 14 & 15 & 16 & 17 & 18 & 19 & 21 & 22 & 23 & 24 & 25 \\
R & 14 & 15 & 16 & 17 & 18 & 19 & 20 & 20 & 21 & 22 & 23 & 24 \\ \hline
L & 26 & 27 & 28 & 29 & 29 & 30 & 31 & 32 & 33 & 34 & 35 & 36 \\
R & 25 & 26 & 27 & 28 & 30 & 31 & 32 & 33 & 34 & 35 & 36 & 37 \\ \hline
L & 38 & 40 & 41 & 42 & 43 & 44 & 45 & 45 & 46 & 47 & 48 & 49 \\
R & 39 & 39 & 40 & 41 & 42 & 43 & 44 & 44 & 45 & 46 & 47 & 48 \\ \hline
\end{tabular}%
}
\vspace{-3mm}
\caption{\small Input images pairs of DTU Dataset}
\label{tab:dtu_pair}
\end{table}

%% file: figures/homo_stable.tex
\begin{figure}[t]
        \centering
        \begin{subfigure}{0.9\linewidth}
            \includegraphics[height=0.16\textheight]{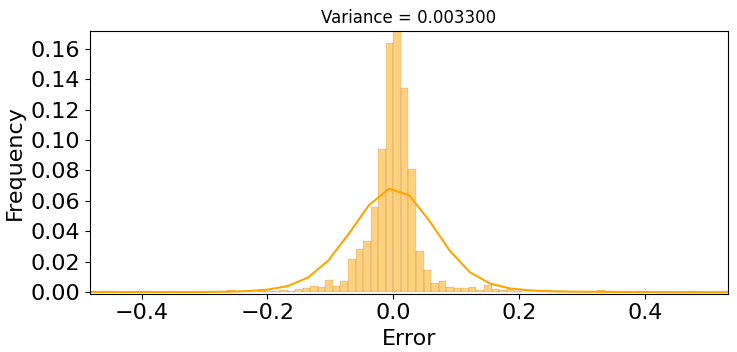}
            \caption{\small \textbf{The statistics of the error in homography estimation.} The curve represents the Gaussian fitting of the statistical results.}
            \label{fig:homo_stable_2}
        \end{subfigure}
        % \vspace{-2mm}
        
        \hfill 
        \vspace{-2mm}

        \begin{subfigure}{0.9\linewidth}
            \centering
            \includegraphics[height=0.16\textheight]{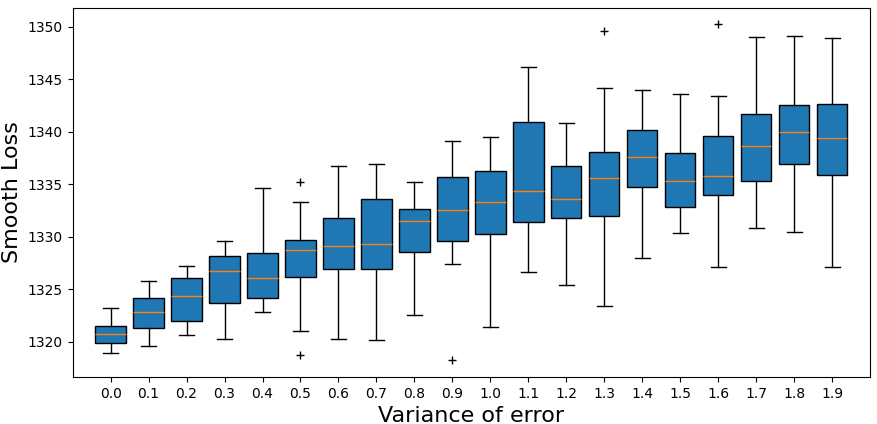}
            \caption{\small \textbf{The simulation of the influence of noise added to the homography matrix.} The blue bars represent the smooth loss error ranges corresponding to noise variances $\sigma$, and the cross points represent outliers.}
            \label{fig:homo_stable_1}
        \end{subfigure}
        \vspace{-2mm}
        \caption{\small The robustness analyze of \homo}
        \label{fig:homo_stable}
    \end{figure}